\relax
\documentclass[11pt,a4paper]{article} %DO NOT CHANGE THIS
\usepackage[hyperref]{naaclhlt2019}  %Required
\usepackage{times}  %Required
\usepackage{latexsym}
\usepackage{url}  %Required
\usepackage{graphicx}  %Required
\newcommand{\dan}[1]{\textcolor{red}{[DR: #1]}}
\newcommand{\danchange}[1]{{\color{blue} {#1}}}
\newcommand{\QN}[1]{#1}

\aclfinalcopy % Uncomment this line for the final submission
\def\aclpaperid{242} %  Enter the acl Paper ID here

\usepackage{url}
\usepackage{xcolor}
\usepackage{amsmath,amsthm}
\usepackage{amssymb,dsfont}
\usepackage{graphicx}
\usepackage{multirow}
\usepackage[linesnumbered,ruled,vlined]{algorithm2e}
\usepackage{algpseudocode}
\usepackage{todonotes}
\usepackage{booktabs,caption}
\usepackage[flushleft]{threeparttable}
\usepackage{textcomp}
\usepackage{subcaption}
\newcommand{\lbl}[1]{\textsc{#1}}
\newcommand{\pad}{\mathcal{P}}
\newcommand{\fad}{\mathcal{T}}
\newcommand{\mcal}[1]{\mathcal{#1}}
\newcommand{\mbf}[1]{\mathbf{#1}}
\newcommand{\event}[1]{\textit{#1}}
\newcommand{\prob}[1]{\text{P}\left(#1\right)}
\newcommand{\ignore}[1]{}
\newtheorem{definition}{Definition}
\newtheorem{theorem}{Theorem}
\newtheorem{example}{Example}
\begin{document}
% The file aaai.sty is the style file for AAAI Press 
% proceedings, working notes, and technical reports.
%
%\title{Partial Or Complete, That Is The Question}
%\title{ \vspace*{-0.5in}
%{{\small \hfill To appear in NAACL'19 (preliminary version)}\\
%\vspace*{.25in}} Partial Or Complete, That Is The Question}
\title{	Partial Or Complete, That Is The Question }
%\ignore{
	%Structured Learning:``Partial or Complete, That is the Question.'' I %delete the data\\
	%``Partial or Complete Structured Data, That is the Question.''\\
	%Partial Or Complete, That Is The Question for Structured Data\\
%}
%\author{Anonymous AAAI Submission}
\author{Qiang Ning,$^1$~Hangfeng He,$^{2}$~Chuchu Fan,$^{1}$~Dan Roth$^{1,2}$\\
$^{1}$Department of Electrical and Computer Engineering, University of Illinois at Urbana-Champaign\\
$^{2}$Department of Computer and Information Science, University of Pennsylvania\\
\texttt{\small \{qning2,cfan10\}@illinois.edu}, \texttt{\small \{hangfeng,danroth\}@seas.upenn.edu}}

%\author{Qiang Ning \and Chuchu Fan\\
%Department of ECE\\
%University of Illinois at Urbana-Champaign\\
%Urbana, IL 61801\\
%\texttt{\small \{qning2,cfan10\}@illinois.edu}
%\And
%Hangfeng He \and Dan Roth\\
%Department of CS\\
%University of Pennsylvania\\
%Philadelphia, PA 19104\\
%\texttt{\small \{hangfeng,danroth\}@seas.upenn.edu}
%}
\maketitle
\begin{abstract}
For many structured learning tasks, the data annotation process is complex and costly. Existing annotation schemes usually aim at acquiring {\em completely} annotated structures, under the common perception that {\em partial} structures are of low quality and could hurt the learning process. This paper questions this common perception, motivated by the fact that structures consist of {\em interdependent} sets of variables. Thus, given a fixed budget, partly annotating each structure may provide the same level of supervision, while allowing for more structures to be annotated. We provide an information theoretic formulation for this perspective and use it, in the context of three diverse structured learning tasks, to show that learning from partial structures can {\em sometimes} outperform learning from complete ones. Our findings may provide important insights into structured data annotation schemes and could support progress in learning protocols for structured tasks.

\ignore{
\dan{A second comment (not relevant to the abstract), given one of Nitish's remarks: I think that somewhere in the introduction we can add a footnote explaining the notion of "cost" -- we need to say that we realize that this is not directly "number" of variables annotated, since the cost may not be uniform; our analysis in fact shows just that. However, in this paper we do not take into account that the time to annotate may also not be uniform. }
}
    
\ignore{
	Collecting supervision signals for machine learning (ML) methods is often costly, especially when the desired annotation is {\em structured}.
	A common strategy is to get supervision from naturally occurring and therefore cheap signals, but one of the issues is that these signals rarely provide {\em complete} structures, for which this paper proposes a general framework to learning from {\em partial} structures.
	Additionally, when intentionally collecting partial annotations, we can spend the annotation budget on more structures, so a natural question is whether learning from complete structures is definitely better than from partial ones.
	This paper first studies this question from an information theory point of view and then shows empirically using three different ML tasks that learning from partial structures can indeed outperform learning from complete ones.
	We hope to share these interesting findings and foster more discussions along this direction.
}
\end{abstract}

% !TEX root = root.tex
\section{Introduction}
\begin{figure}[h!]
	\centering
	\includegraphics[width=.5\textwidth]{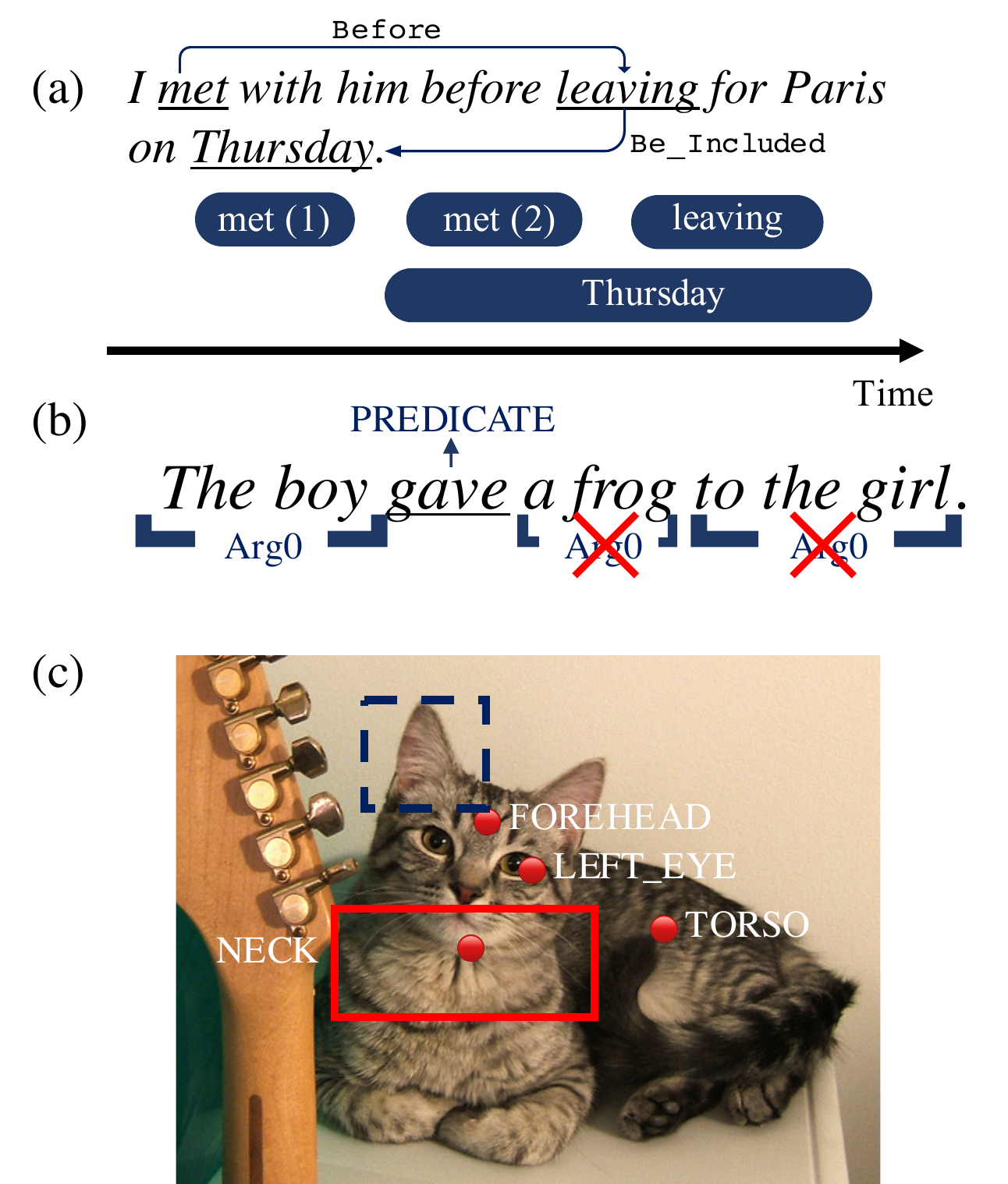}
	\caption{Due to the inherent structural constraints of each task, individual instances therein put restrictions on others. (a) The temporal relation between \event{met} and \event{Thursday} has to be \lbl{Before} (``met (1)'') or \lbl{Be\_Included} (``met (2)''). (b) The argument roles of {\em a frog} and {\em to the girl} cannot be \lbl{Arg0} anymore. (c) Given the position of the cat's FOREHEAD and LEFT\_EYE, a rough estimate of its NECK can be the red solid box rather than the blue dashed box.}
	\label{fig:closure not working}
\end{figure}
Many machine learning tasks require structured outputs, and the goal is to assign values to a set of variables coherently. 
Specifically, the variables in a structure need to satisfy some global properties required by the task.
An important implication is that once some variables are determined, the values taken by other variables are constrained.
%if some variables are already determined, then other variables will be constrained as well.
For instance, in the temporal relation extraction problem in  Fig.~\ref{fig:closure not working}a, if \event{met} happened before \event{leaving} and \event{leaving} happened on \event{Thursday}, then 
%without referring to the text, 
we know that \event{met} must either be before \event{Thursday} (``met (1)'') or has to
%also 
happen on \event{Thursday}, too (``met (2)'') \cite{NFWR18}.
Similarly, in the semantic frame of the predicate {\em gave} \cite{KingsburyPa02} in Fig.~\ref{fig:closure not working}b, if {\em the boy} is \lbl{Arg0} (short for argument 0), then it rules out the possibility of {\em a frog} or {\em to the girl} taking the same role.
%g \lbl{Arg0} again.
Figure~\ref{fig:closure not working}c further shows an example of part-labeling of images
\cite{CKKF18}; given the position of FOREHEAD and LEFT\_EYE of the cat in the picture, we roughly know that its NECK should be somewhere in the red solid box, while the blue dashed box is likely to be wrong.

\begin{figure}[htbp!]
	\centering
	\includegraphics[width=.45\textwidth]{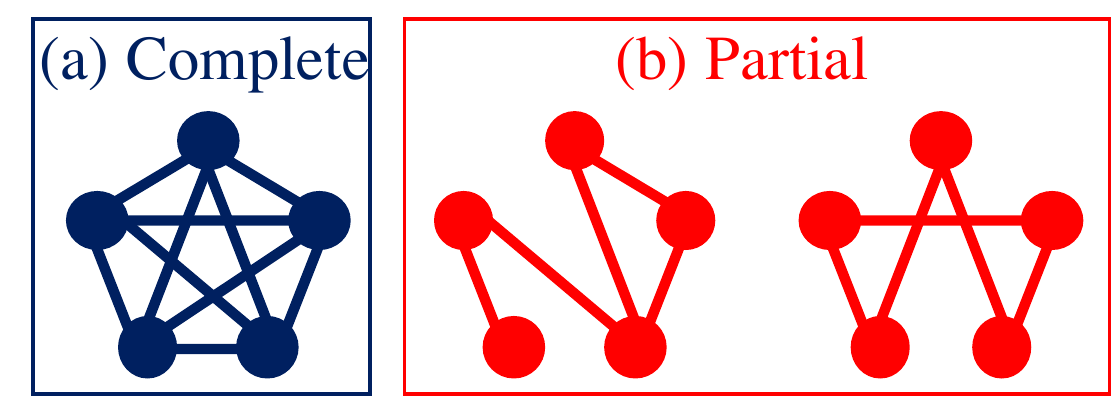}
	\caption{If we need training data for a graph labeling task (assuming the gold values for the nodes are given) and our annotation budget allows us to annotate, for instance, 10 edges in total, we could (a) completely annotate one graph (and then we run out of budget), or (b) partially annotate two graphs.}
	\label{fig:complete vs full}
\end{figure}

Data annotation for these structured tasks is complex and costly, thus requiring one to make the most of a given budget.
{This issue has been investigated for decades from the perspective of active learning for classification tasks \cite{Angluin88a,AtlasCoLa90,LewisGa94a} and for structured tasks \cite{RothSm06a,RothSm06,RothSm08,HLAR19}.
% \citet{RothSm06a,RothSm06,RothSm08} extended conventional active learning to structured tasks via margin-based querying functions, pipeline models, and interactive feature construction.
While active learning aims at selecting the next structure to label, we try to investigate, from a different perspective, whether we should annotate each structure completely or partially.}
Conventional annotation schemes typically require {\em complete} structures, under the common perception that partial annotation could adversely affect the performance of the learning algorithm.
%{\em partialness} is of low quality for learning algorithms and should be minimized.
But note that partial annotations will allow for more structures to be annotated (see Fig.~\ref{fig:complete vs full}).
Therefore, a fair comparison should 
be done while maintaining a fixed annotation budget, which was not done before.
%take into account the size difference, 
%which has hardly been investigated before. 
Moreover, even if partial annotation leads to comparable learning performance to conventional complete schemes, it provides
%we will have 
more flexibility in data annotation.
% For example, if the task is to label the edges in graphs with 5 nodes and we are limited to label 10 edges, then only one graph can be completed (Fig.~\ref{fig:complete vs full}a); in contrast, labeling only 5 edges in each graph can instead give us two graphs (Fig.~\ref{fig:complete vs full}b).
% While we agree that completeness is necessary for {\em test} data, we question whether this is also the case for {\em training} data, because partial annotations will allow for more structured to be annotated under the same budget, as shown by Fig.~\ref{fig:complete vs full}b.

Another potential benefit of %partialness 
partial annotation is that it imposes constraints on the remaining parts of a structure. 
As illustrated by Fig.~\ref{fig:closure not working}, with partial annotations, we 
%will still 
already have some knowledge about the unannotated parts.
Therefore, further annotations of these variables may use the available budget less efficiently;
%on these parts may be less efficient in making use of a given budget; 
this effect was first discussed in \citet{NYFR18}.
Motivated by the observations in Figs.~\ref{fig:closure not working}-\ref{fig:complete vs full}, we think it is important to study partialness systematically, before we hastily assume that {\em completeness} should always be favored in data collection.

% which poses a new question to us: Given a fixed budget, is complete possibly better if we spend the annotation resources on a new structure other than the current structure that we already know of?
% if complete annotations are required (as by conventional schemes), an annotator needs to compare \event{met} and \event{Thursday} temporally, label the argument roles of {\em a frog} and {\em to the girl}, and mark the exact position of the cat's neck, respectively.
% An issue with that is the efficiency of obtaining as much information as possible given a fixed budget, which poses a new question to the community: Why not spend the precious annotation resources on a new structure other than the current structure that we already know of (partially)?
%In the aforementioned annotations, the facts that \event{met} is \lbl{before} \event{Thursday}, or that {\em a frog} is \lbl{Arg1}, are less surprising when we already have some knowledge about the labels of other parts of the structure.
%This raises a key question: Why not spend the precious annotation resources on components of other structures, those that are less constrained by existing annotations?

{To study whether the above benefits of partialness can offset its weakness for learning, \textbf{our first contribution} is the proposal of \textit{early stopping partial annotation} (ESPA)} scheme, which randomly picks up instances to label in the beginning, and stops before a structure is completed. \QN{We do not claim that ESPA should {\em always} be preferred; instead, it serves as an {\em alternative} to %the 
conventional, complete annotation schemes that we should keep in mind, because, as we show later, it can be comparable to (and sometimes even better than) complete annotation schemes.}
%{\bf Our first contribution} is that we propose an {\em early stopping partial annotation} (ESPA) scheme as an alternative to , which initially performs annotation as in conventional schemes, but stops before a structure is completed, saving the budget for more structures (Fig.~\ref{fig:complete vs full}b).
ESPA is straightforward to implement even in crowdsourcing; instances to annotate can be selected {\em offline} and distributed to crowdsourcers; this can be contrasted with the difficulties of implementing active learning protocols in these settings \cite{AmbatiVoCa10,LawsScSc11}.
We think that ESPA is a good representative for a systematic study of partialness.
% ESPA is also compatible with active learning -- just stop early and move on to the next structure before completing the current one.

% ESPA presents an interesting trade-off for structured learning: a few {\em complete}  structures vs. many {\em partial} structures.
{\bf Our second contribution} is the development of an information theoretic formulation to explain the benefit of ESPA (Sec.~\ref{sec:early stop}), which we further demonstrate via three structured learning tasks in Sec.~\ref{sec:exp}: temporal relation (TempRel) extraction \cite{ULADVP13}, semantic role classification (SRC),\footnote{A subtask of semantic role labeling (SRL) \cite{PalmerGiXu10} that only classifies the role of an argument.} and shallow parsing \cite{TjongBu00}.
\QN{These tasks are chosen because they each represent a wide spectrum of structures that we will detail later.}
\textbf{As a byproduct}, we extend constraint-driven learning (CoDL) \cite{ChangRaRo07} to cope with partially annotated structures (Sec.~\ref{sec:algorithm});  we call the algorithm {\em Structured Self-learning with Partial ANnotations} (SSPAN) to distinguish it from CoDL.\footnote{There has been many works on learning from partial annotations, which we review in Sec.~\ref{sec:algorithm}. SSPAN is only an experimental choice in demonstrating ESPA.
Whether SSPAN is better than other algorithms is out of the scope here, and a better algorithm for ESPA will only strengthen the claims in this paper.}

We believe in the importance of work in this direction.
{\em First, partialness is inevitable in practice}, either by mistake or by choice, so
%\footnote{\dan{why this example?}To rank a gigantic database with millions of ads, friends, news feeds, etc., we often have to choose partial annotations.} 
our theoretical analysis can provide unique insight into understanding partialness.
{\em Second, it opens up opportunities for new annotation schemes.} Instead of considering partial annotations as a compromise,
%when completeness is infeasible, 
we can in fact annotate partial data {\em intentionally}, allowing us to design favorable guidelines and collect more important annotations at a cheaper price. Many recent datasets that were collected via crowdsourcing are already partial, and this paper provides some theoretical foundations for them. Furthermore, the setting described here addresses natural scenarios where only partial, indirect supervision is available, as in {\em Incidental Supervision} \citep{Roth17}, and this paper begins to provide theoretical understanding for this paradigm, too.
Further discussions can be found in Sec.~\ref{sec:discussion}.

\ignore{
{An important assumption here is that the cost incurred by each individual annotation is the same (that is, all edges in Fig.~\ref{fig:complete vs full} cost equally), often the default setting in crowdsourcing. In other cases, however, the difficulty and cost can vary a lot for different annotation instances, so we randomly select instances to label to guarantee that the assumption holds on average.
Moreover, the cost of a piece of annotation may be correlated with how informative that annotation is, but we leave it for future studies such as active learning schemes.
For fairness, we randomly select instances to label so that on average, our assumption still holds.}
\danchange{It is important to clarify that we assume uniform cost over individual annotations (that is, all edges in Fig.~\ref{fig:complete vs full} cost equally), often the default setting in crowdsourcing. However, the annotation difficulty can vary a lot in some cases. Moreover, we realize that the cost of a piece of annotation may be correlated with how informative it is. We leave this issue for future studies, possibly in the context of active learning schemes. For fairness, we randomly select instances to be labeled so that on average, our assumption still holds.}
}

It is important to clarify that we assume uniform cost over individual annotations (that is, all edges in Fig.~\ref{fig:complete vs full} cost equally), often the default setting in crowdsourcing. \QN{We realize that the annotation difficulty can vary a lot in practice, sometimes incurring different costs. To address this issue, we randomly select instances to label so that on average, the cost is uniform. We agree that, even with this randomness, there could still be situations where the assumption does not hold, but we leave it for future studies, possibly in the context of active learning schemes.}
% !TEX root = root.tex
\section{ESPA: Early Stopping Partial Annotation}
\label{sec:early stop}
In this section, we study whether the effect demonstrated by the examples in Fig.~\ref{fig:closure not working} exists in general. 
First, we formally define {\em structure} and {\em annotation}. 
% Note that we use {\em instances} to denote variables of a structure that one needs to label.
\begin{definition}
	A structure of size $d$ is a vector of random variables (RV) $\mbf{Y}=[Y_1,\dots,Y_d]\in C(\mcal{L}^d)$, where $\mcal{L}=\{\ell_1,\dots,\ell_{|\mcal{L}|}\}$ is the label set for each variable and $C(\mcal{L}^d)\subseteq\mcal{L}^d$ represents the constraints imposed by this type of structure.
\end{definition}
It is necessary to model a structure as a set of {\em random} variables because when it is not completely annotated, there is still uncertainty in the annotation assignment. To study partial annotations, we introduce the following:
% Annotations are essentially reducing this uncertainty by labeling its variables:
\begin{definition}
	A $k$-step annotation ($0\le k\le d$) is a vector of RVs $\mbf{A}_k=[A_{k,1},\dots,A_{k,d}]\in \left(\mcal{L}\cup \sqcap\right)^d$ where $\sqcap$ is a special character for null, such that
	\begin{equation}
	\sum_{i=1}^{d}{\mathds{1}(A_{k,i}\ne\sqcap)} = k,
	\label{eq:k annotated}
	\end{equation}
	\QN{\begin{equation}
	\prob{\mbf{Y}|\mbf{A}_k=\mbf{a}_k}=\prob{\mbf{Y}|{Y}_{j}={a}_{k,j}, j\in\mathcal{J}},
% 	\prob{Y_i | A_{k,i}} = \begin{cases}
% 	\mathds{1}(Y_i=A_{k,i}) & A_{k,i}\ne \sqcap\\
% 	\prob{Y_i} & A_{k,i}=\sqcap
% 	\end{cases},
	\label{eq:annotated}
	\end{equation}}
% 	where $i=1,\dots,d$.
    \QN{where $\mathcal{J}$ is the set of indices that $a_{k,j}\ne\sqcap$.}
	\ignore{
	\begin{equation}
		\prob{\mbf{Y}=\mbf{y},\mbf{A}_k=\mbf{a}_k} = \prob{\mbf{Y}=\mbf{y}^\prime},
	\end{equation}
	where $y_i^\prime = \begin{cases}
	y_i & a_{k,i} = \sqcap\\
	a_{k,i} & a_{k,i}\ne \sqcap
	\end{cases}$, $i=1,\dots,d$.}
\end{definition}
Eq.~\eqref{eq:k annotated} means that, in total, $k$ variables are already annotated at step $k$. 
Obviously, $\mbf{A}_0$ means that no variables are labeled, and $\mbf{A}_d$ means that all variables in $\mbf{Y}$ are determined. $\mbf{A}_k$ is what we call a $k$-step ESPA, so hereafter we use $k/d$ to represent annotation completeness.
Eq.~\eqref{eq:annotated} assumes no annotation mistakes, so if the $i$-th variable is labeled, then $Y_i$ must be the same as $A_{k,i}$. 
\ignore{Based on Eq.~\eqref{eq:annotated}, we also have
\begin{equation}
\prob{\mbf{Y}|\mbf{A}}=\prob{\mbf{Y}|{Y}_{j}={A}_{k,j}, j\in\mathcal{J}},
\label{eq:joint condition prob}
\end{equation}
where $\mathcal{J}$ is the set of indices that $A_{k,j}\ne\sqcap$, which we will find useful next.}

To measure the theoretical benefit of $\mbf{A}_k$, we propose the following quantity
\begin{equation}
	I_k=\log{|C(\mcal{L}^d)|}-E\left[\log{f(\mbf{a}_k)}\right]
	\label{eq:information in structure}
\end{equation}
for $k=0,\dots,d$, where $f(\mbf{a}_k)=\lvert\{\mbf{y}\in C(\mcal{L}^d): \prob{\mbf{y}|\mbf{a}_k}> 0\}\rvert$ is the total number of structures in $C(\mcal{L}^d)$ that are still valid given $\mbf{A}_k=\mbf{a}_k$.
Since we assume that the labeled variables in $\mbf{A}_k$ are selected uniformly randomly, $E\left[\cdot\right]$ is simply the average of $\log{f(\mbf{a}_k)}$.
When $k=0$, $f(\mbf{a}_k)\equiv C(\mcal{L}^d)$ and $I_0\equiv 0$; as $k$ increases,
$I_k$ increases since the structure has more and more variables labeled; finally, when $k=d$, the structure is fully determined and $I_d\equiv \log{|C(\mcal{L}^d)|}$.
The first-order finite difference, $I_k-I_{k-1}$, is the benefit brought by annotating an additional variable at step $k$;
if $I_k$ is concave (i.e., a decaying $I_k-I_{k-1}$),
% \footnote{A sequence of real numbers \{$a_n$\}, $n=0,1,\dots$ is concave, if it satisfies $a_n-a_{n-1}\le a_{n+1}-a_n$, $n=1,2,\dots$.} 
the benefit from a new annotation attenuates, suggesting the potential benefit of the ESPA strategy.

In an extreme case where the structure is so strong that it requires all individual variables to share the same label, then labeling any variable is sufficient for determining the entire structure. Intuitively, we do not need to annotate more than one variable.
Our $I_k$ quantity can support this intuition: The structural constraint, $C(\mcal{L}^d)$, contains only $|\mcal{L}|$ elements: $\{[\ell_i,\ell_i,\dots,\ell_i]\}_{i=1}^{|\mcal{L}|}$, so $I_0=0$, and $I_1=\cdots=I_d=\log{|\mcal{L}|}$. Since $I_k$ does not increase at all when $k>=1$, we should adopt first-step annotation $\mbf{A}_1$.
Another extreme case is that of a trivial structure that has no constraints (i.e., $C(\mcal{Y}^d)=\mcal{Y}^d$). The annotation of all variables are independent and we gain no advantage from skipping any variables.
This intuition can be supported by our $I_k$ analysis as well: Since $I_k=k\log{|\mcal{L}|}$, $\forall k=0,1,\dots,d$, $I_k$ is linear and all steps contribute equally to improving $I_k$ by $\log{|\mcal{L}|}$; therefore ESPA is not necessary.

Real-world structures are often not as trivial as the two extreme cases above, but $I_k$ can still serve as a guideline to help determine whether it is beneficial to use ESPA. We next discuss three diverse types of structures and how to obtain $I_k$ for them.

\begin{figure}[h!]
	\centering
	\includegraphics[width=.5\textwidth]{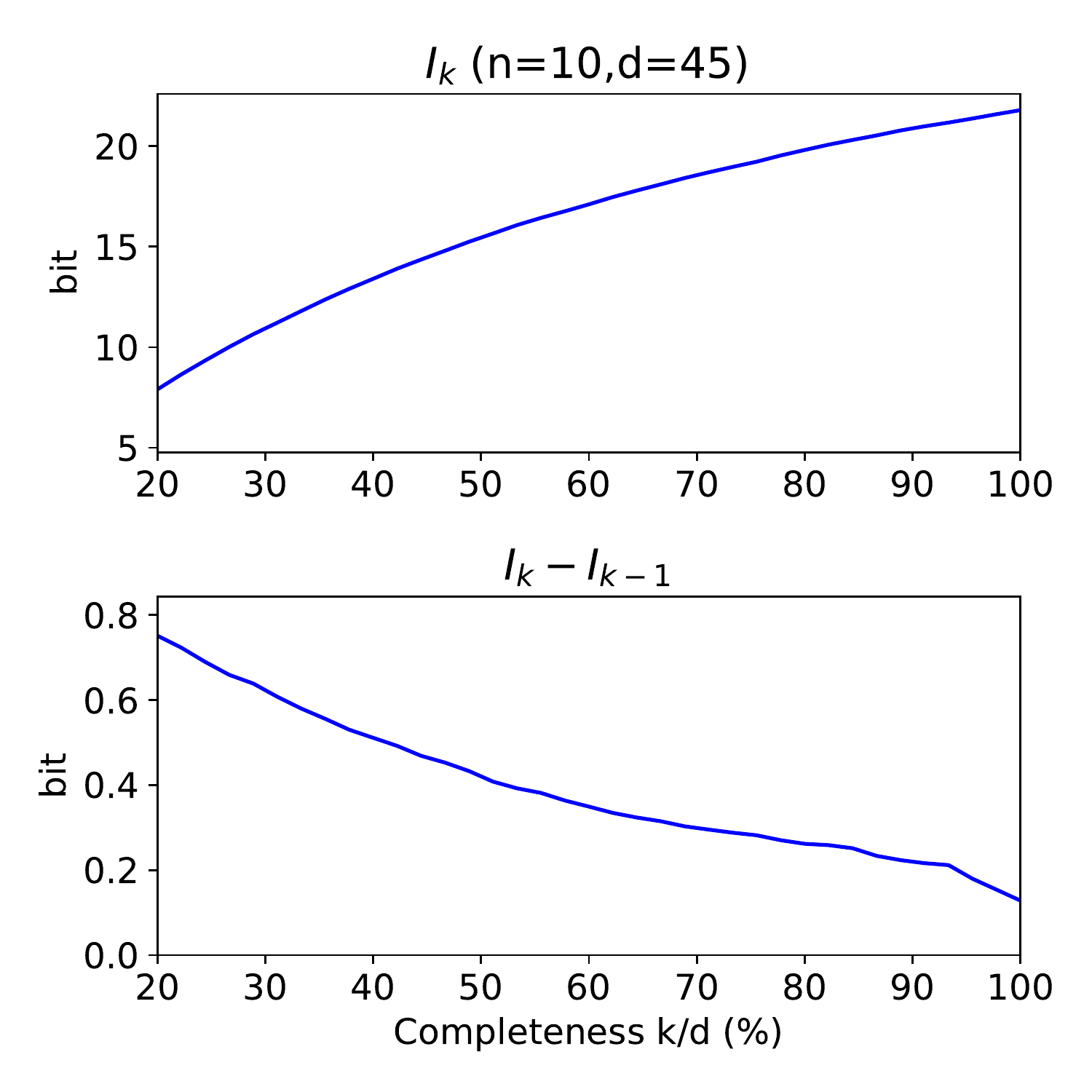}
	\caption{The mutual information between the chain structure and its $k$-step ESPA, $I_k$, is concave, suggesting possible benefit of using ESPA. In the simulation, there are $n=10$ items in the chain and thus $d=45$ pairs, $k$ of which are labeled. The values of $I_k$'s, as defined by Eq.~\eqref{eq:information in structure}, were obtained through averaging 1000 experiments. \QN{We use base-2 logarithm and the unit on y-axis is thus ``bit''.}}
	\label{fig:topo sort}
\end{figure}

\begin{example}
	\label{ex:dag}
	The ranking problem is an important machine learning task and often depends on pairwise comparisons, for which the label set is $\mcal{L}=\{<,>\}$. For a ranking problem with $n$ items, there are $d=n(n-1)/2$ pairwise comparisons in total. Its structure is a chain following the transitivity constraints, i.e., if $A < B$ and $B < C$, then $A < C$.
\end{example}
A $k$-step ESPA $\mbf{A}_k$ for a chain means that only $k$ (out of $d$) pairs are compared and labeled, resulting in a directed acyclic graph (DAG).
In this case, $f(\mbf{a}_k)$ is actually counting the number of linear extensions of the DAG, which is known to be \#P-complete \cite{BrightwellWi91}, so we do not have a closed-form solution to $I_k$. In practice, however, we can use the Kahn's algorithm and backtracking to simulate with a relatively small $n$, as shown by Fig.~\ref{fig:topo sort}, where $n=10$ and $I_k$ was obtained through averaging 1000 random simulations. $I_k$ is concave, as reflected by the downward shape of $I_k-I_{k-1}$.
Therefore, new annotations are less and less efficient for the chain structure, suggesting the usage of ESPA.

\begin{figure}[h!]
	\centering
	\includegraphics[width=.5\textwidth]{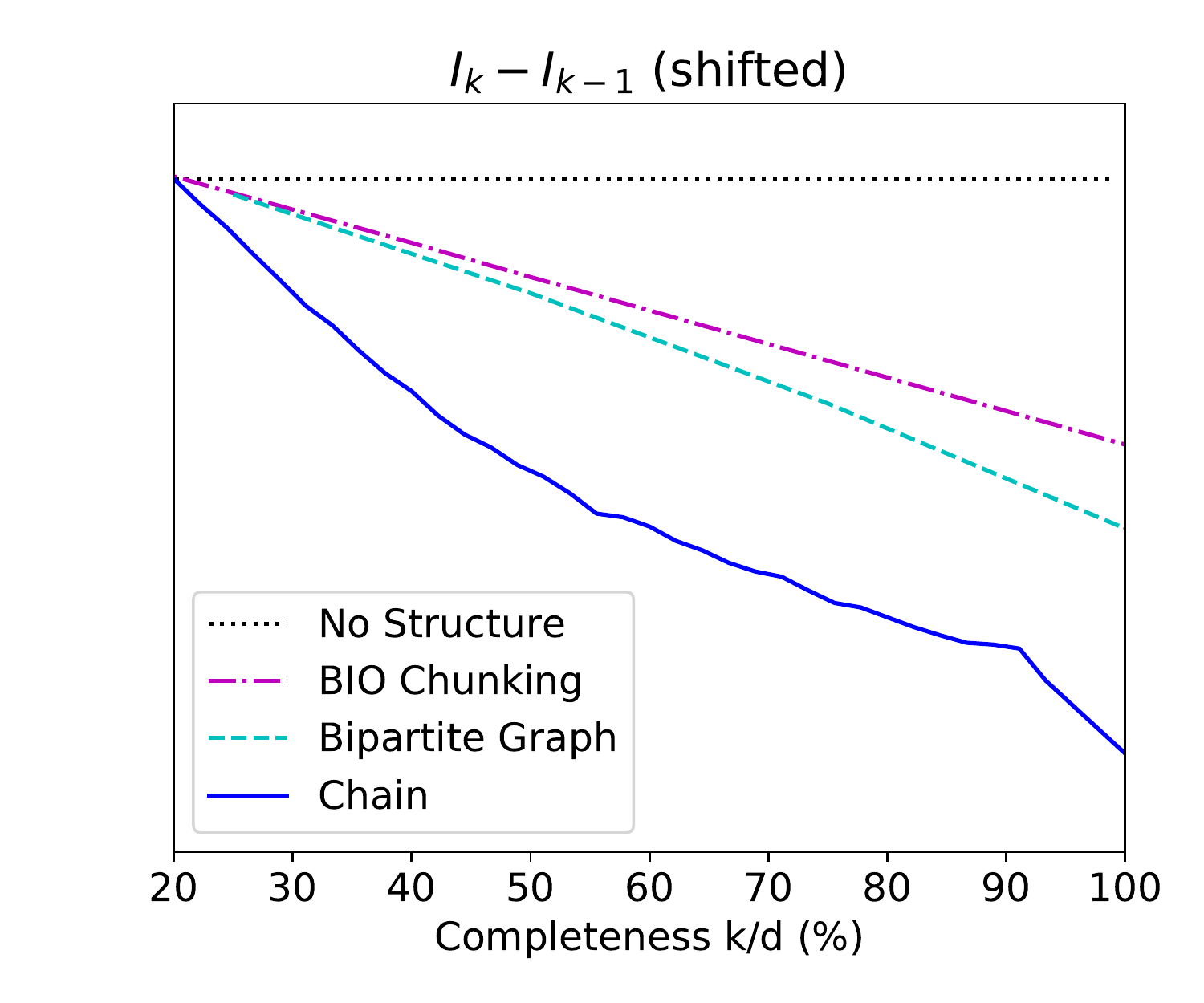}
	\caption{The $I_k-I_{k-1}$ curves from several different structures. The curves are shifted to almost the same starting point for better visualization, so the Y-Axis grid is not shown. \QN{The curve for ``Chain'' was obtained via simulations, and the other curves all have closed-form formulations.}}
	\label{fig:compare convexity}
\end{figure}

\begin{example}
	\label{ex:srl}
	The general assignment problem requires assigning $d$ agents to $d^\prime$ tasks such that the agent nodes and the task nodes form a bipartite graph (without loss of generality, assume $d\le d^\prime$). That is, an agent can handle exactly one task, and each task can only be handled by at most one agent. Then from the agents' point of view, the label set for each of them is $\mcal{L}=\{1,2,\dots,d^\prime\}$, denoting the task assigned to the agent. 
	% Note that the classic M-ary classification problem is a special case with $d=1$ and $d^\prime=M$.
\end{example}
A $k$-step ESPA $\mbf{A}_k$ for this problem means that $k$ agents are already assigned with tasks, and $f(\mbf{a}_k)$ is to count the valid assignments of the remaining tasks to the remaining $d-k$ agents, to which we have closed-form solutions: $f(\mbf{a}_k)=\frac{(d^\prime-k)!}{(d^\prime-d)!}$, $\forall \mbf{a}_k$. According to Eq.~\eqref{eq:information in structure}, $I_k = \log{\frac{d^\prime!}{(d^\prime-k)!}}$ regardless of $d$ or the distribution of $\mbf{A}_k$, and is concave (Fig.~\ref{fig:compare convexity} shows an example of it when $d=4, d^\prime=10$).
\ignore{
$$
I_k-I_{k-1} =\log\left(d^\prime-k+1\right),~1\le k \le d \le d^\prime
$$
}

\begin{example}
	\label{ex:chunking}
	Sequence tagging is an important NLP problem, where the tags of tokens are interdependent. Take chunking as an example. A basic scheme is for each token to choose from three labels, B(egin), I(nside), and O(utside), to represent text chunks in a sentence. That is, $\mcal{L}=\{B,I,O\}$. Obviously, O cannot be immediately followed by I.
\end{example}
Let $d$ be the number of tokens in a sentence.
A $k$-step ESPA $\mbf{A}_k$ for chunking means that $k$ tokens are already labeled by B/I/O, and $f(\mbf{a}_k)$ counts the valid BIO sequences that do not violate those existing annotations. Again, as far as we know, there is no closed-form solution to $f(\mbf{a}_k)$ and $I_k$, but in practice, we can use dynamic programming to obtain $f(\mbf{a}_k)$ and then $I_k$ using Eq.~\eqref{eq:information in structure}. We set $d=10$ and show $I_k-I_{k-1}$ for this task in Fig.~\ref{fig:compare convexity}, where we observe the same effect we see in previous examples: The benefit provided by labeling a new token in the structure attenuates.

Interestingly, based on Fig.~\ref{fig:compare convexity}, we find that \textbf{the slope of $I_k-I_{k-1}$ may be a good measure of the 
``tightness'' or ``strength''
of a structure}. When there is no structure at all, the curve is flat (black). 
The BIO structure is intuitively simple, and it indeed has the flattest slope among the three structured tasks (purple).
When the structure is a chain, the level of uncertainty goes down rapidly with every single annotation (think of standard sorting algorithms); the constraint is intuitively strong and in Fig.~\ref{fig:compare convexity}, it indeed has a steep slope (blue).

\textbf{Finally, we want to emphasize that the definition of $I_k$ in Eq.~\eqref{eq:information in structure} is in fact backed by information theory.}
When we do not have prior information about $\mbf{Y}$, we can assume that $\mbf{Y}$ follows a uniform distribution over $C(\mcal{L}^d)$. Then, $I_k$ is essentially the mutual information between structure $\mbf{Y}$ and annotation $\mbf{A}_k$, $I(\mbf{Y};\mbf{A}_k)$:
\begin{align*}
I(\mbf{Y};\mbf{A}_k) &= H(\mbf{Y}) - H(\mbf{Y}|\mbf{A}_k)\\
&=\log{|C(\mcal{L}^d)|}-E\left[H(\mbf{Y}|\mbf{A}_k=\mbf{a}_k)\right]\\
&=\log{|C(\mcal{L}^d)|}-E\left[\log{f(\mbf{a}_k)}\right],
\end{align*}
where $H(\cdot)$ is the entropy function.
This is an important discovery, since it points out a new way to view a structure and its annotations. It may be useful for studying active learning methods for structured tasks, and other annotation phenomena such as noisy annotations.
\QN{The usage of mutual information also aligns well with the information bottleneck framework \cite{ShamirSaTi10,ShwartzTi17,YuPr18}, although a more recent paper challenges the interpretation of information bottleneck \cite{SBDAKTC18}.}

% !TEX root = root.tex
\section{Learning {from} Partial Structures}
\label{sec:algorithm}
So far, we have been advocating the ESPA strategy to maximize the information we can get from a fixed budget.
Since early stopping leads to partial annotations, one missing component before we can benefit from it is an approach to learning from partial structures.
In this study, we assume the existence of a relatively small but complete dataset that can provide a good initialization for learning from a partial dataset, which is very similar to semi-supervised learning (SSL).
SSL, in its most standard form, studies the combined usage of a labeled set $\mcal{T}=\{(x_i,y_i)\}_i$ and an unlabeled set $\mcal{U}=\{x_j\}_j$, where the $x$'s are instances and $y$'s are the corresponding labels.
SSL gains information about $p(x)$ through $\mcal{U}$, which may improve the estimation of $p(y|x)$. %\cite{ChapelleScZi06}. 
Specific algorithms range from self-training \cite{Scudder65,Yarowsky95}, co-training \cite{BlumMi98},  generative models \cite{NMTM00}, to transductive SVM \cite{Joachims99} etc., among which one of the most basic algorithms is Expectation-Maximization (EM) \cite{DempsterLaRu77}. 
By treating them as hidden variables, EM ``marginalizes'' out the missing labels of $\mcal{U}$ via expectation (i.e., soft EM) or maximization (i.e., hard EM).
For structured ML tasks, soft and hard EMs turn into posterior regularization (PR) \cite{GGGT10} and constraint-driven learning (CoDL) \cite{ChangRaRo07}, respectively.
%All these variants of EM can be derived from the unified expectation maximization (UEM) framework \cite{SamdaniChRo12a}.

Unlike unlabeled data, the partially annotated structures caused by early stopping urge us to gain information not only about $p(x)$, but also from their labeled parts.
There have been many existing work along this line \cite{TKOMM08,FernandesBr11,HovyHo12,LouHa12}, but in this paper, we decide to extend CoDL to cope with partial annotations due to two reasons. First, CoDL, which itself can be viewed as an extension of self-training to {\em structured} learning, is a wrapper algorithm having wide applications. Second, as its name suggests, CoDL learns from $\mcal{U}$ by guidance of constraints, so partial annotations in $\mcal{U}$ are technically easy to be added as extra equality constraints.

Algorithm~\ref{algo:sspan} describes our {\em Structured Self-learning with Partial ANnotations} (SSPAN) algorithm \QN{that learns a model $\mathcal{H}$}. The same as CoDL, SSPAN is a wrapper algorithm requiring two components: \textsc{Learn} and \textsc{Inference}. \textsc{Learn} attempts to estimate the {\em local} decision function for each individual instance regardless of the {\em global} constraints, while \textsc{Inference} takes those local decisions and performs a {\em global} inference.
Lines~\ref{ln:self training start}-\ref{ln:self training end} are the procedure of self-training, which iteratively completes the missing annotations in $\mcal{P}$ and learns from both $\mcal{T}$ and the completed version of $\mcal{P}$ (i.e., $\tilde{\mcal{P}}$).\footnote{Line~\ref{ln:self training end} can be interpreted in different ways, either as $\mcal{T}\cup\tilde{\mcal{P}}$ (adopted in this work) or as a weighted combination of 
	\textsc{Learn}($\mcal{T}$) and \textsc{Learn}($\tilde{\mcal{P}}$) (adopted by \cite{ChangRaRo07}). }
Line~\ref{ln:structure constraints} requires that the inference follows the structural constraints inherently in the task, turning the algorithm into CoDL; Line~\ref{ln:partial constraints} enforces those partial annotations in $\mbf{a}_i$,  further turning it into SSPAN.
In practice, \textsc{Inference} can be realized by the Viterbi or beam search algorithm in sequence tagging, or more generally, by Integer Linear Programming (ILP) \cite{PRYZ05}; either way, the partial constraints of Line~\ref{ln:partial constraints} can be easily incorporated.

\begin{algorithm}
	\DontPrintSemicolon % Some LaTeX compilers require you to use \dontprintsemicolon instead
	\KwIn{\small $\fad{}=\{(\mbf{x}_i,\mbf{y}_i)\}_{i=1}^N$, $\pad{}=\{(\mbf{x}_i,\mbf{a}_i)\}_{i=N+1}^{N+M}$}
	Initialize $\mcal{H}=\textsc{Learn}(\fad{})$ \label{ln:init}\;
	\While{convergence criteria not satisfied}{
		$\tilde{\mathcal{P}}=\emptyset$\label{ln:self training start}\;
		\ForEach{$(\mbf{x}_i,\mbf{a}_i)\in\mathcal{P}$}{
			$\hat{\mathbf{y}}_i$ = \textsc{Inference}($\mbf{x}_i;\mcal{H}$)\label{ln:inference}, such that\;
			\hspace{3em}$\diamond$~~$\hat{\mbf{y}}_i\in C(\mcal{Y}^d)$ \label{ln:structure constraints}\;
			\hspace{3em}$\diamond$~~$\hat{y}_{i,j}=a_{i,j}$, $\forall a_{i,j}\ne \sqcap$ \label{ln:partial constraints}\;
			$\tilde{\mathcal{P}}=\tilde{\mathcal{P}}\cup \{(\mathbf{x}_i,\hat{\mathbf{y}}_i)\}$\;
		}
		$\mcal{H} =\textsc{Learn}(\fad+\tilde{\mathcal{P}})$\label{ln:learn2}\label{ln:self training end}\;
	}
	\Return{$\mcal{H}$}\;
	\caption{Structured Self-learning with Partial Annotations (SSPAN)}
	\label{algo:sspan}
\end{algorithm}
% !TEX root = root.tex
\section{Experiment}
\label{sec:exp}

%Based on SSPAN, we can investigate the effect of early stopping partial annotations.
In Sec.~\ref{sec:early stop}, we argued from an information theoretic view that ESPA is beneficial for structured tasks if we have a fixed annotation resource.
We then proposed SSPAN in Sec.~\ref{sec:algorithm} to learn from the resulting partial structures.
\textbf{However}, on one hand, there is still a gap between the $I_k$ analysis and the actual system performance; on the other hand, whether the benefit can be realized in practice also depends on how effective the algorithm exploits partial annotations.
Therefore, it remains to be seen how ESPA works in practice.
Here we use three NLP tasks: temporal relation (TempRel) extraction, semantic role classification (SRC), and shallow parsing, analogous to the chain, assignment, and BIO structures, respectively.

\begin{figure}[h!]
	\centering
	\includegraphics[width=.5\textwidth]{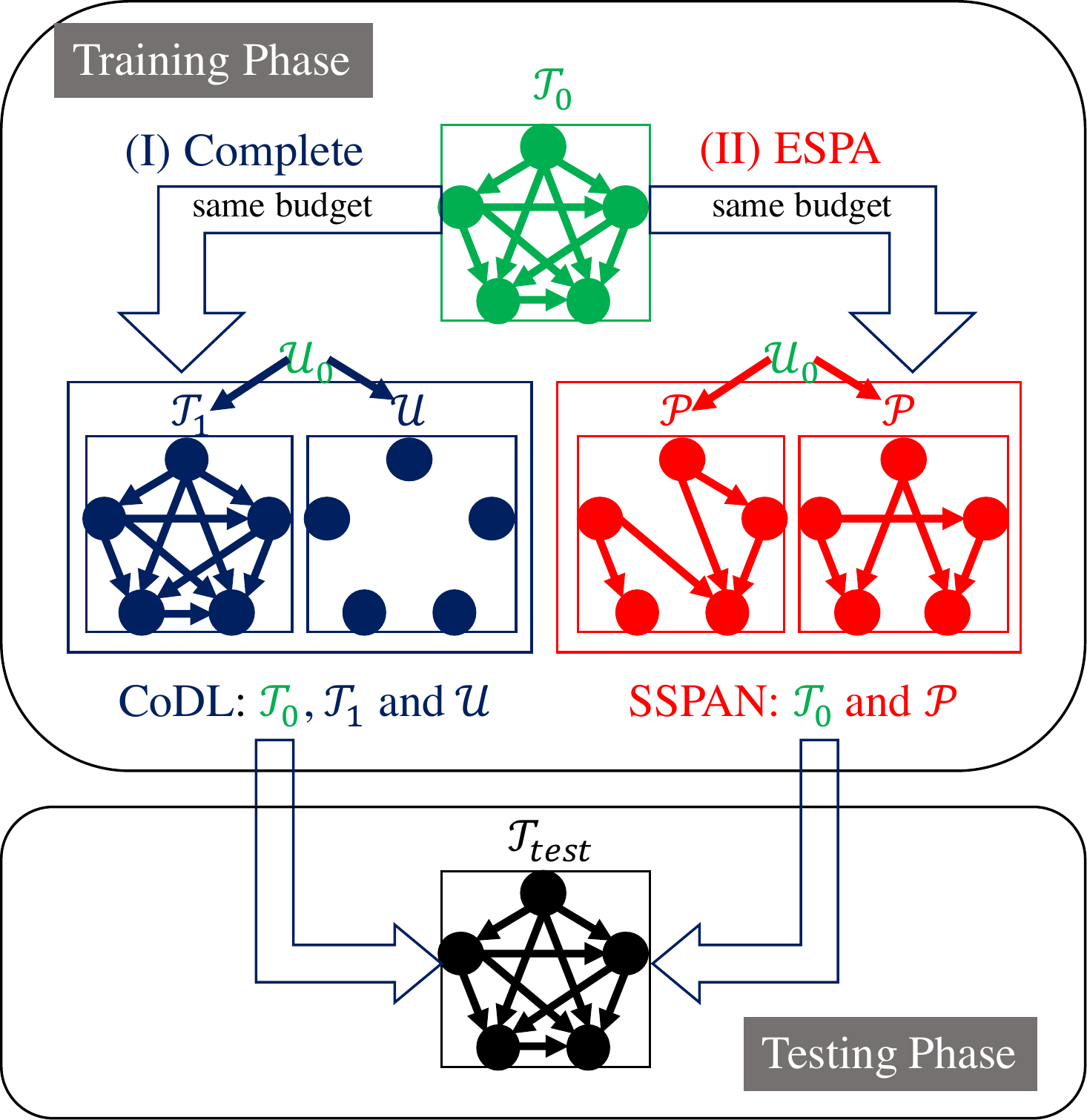}
	\caption{The two annotation schemes we compare in Sec.~\ref{sec:exp}. $\mcal{T}$, $\pad$, and $\mcal{U}$ denote complete, partial, and empty structures, respectively. Both schemes start with a complete and relatively small dataset and an unannotated dataset (green). (I) Conventional complete annotation scheme (blue). (II) The proposed ESPA scheme (red). Finally, they are tested on an unseen and complete dataset (black).}
	\label{fig:comparison scheme}
\end{figure}

For all tasks, we compare the following two schemes in Fig.~\ref{fig:comparison scheme}, where we use graph structures for demonstration.
Initially, we have a relatively small but complete dataset $\mcal{T}_0$, an unannotated dataset $\mcal{U}_0$, and some budget to annotate $\mcal{U}_0$. The conventional scheme I, also our baseline here, is to annotate each structure completely before randomly picking up the next one. Due to the limited budget, some $\mcal{U}_0$ remain untouched (denoted by $\mcal{U}$). The proposed scheme II adopts ESPA so that all structures at hand are annotated but only partially.
For fair comparisons, we use CoDL to incorporate $\mcal{U}$ into scheme I as well. 
Finally, the systems trained on the dataset from I/II via CoDL/SSPAN are evaluated on unseen but complete testset $\mcal{T}_{test}$.
Note that because ESPA is a new annotation scheme, there exists no dataset collected this way. We use existing complete datasets and randomly throw out some annotations to mimic ESPA in the following.
Due to the randomness in selecting which structures/instances to keep in scheme I/II, we repeat the whole process multiple times and report the mean $F_1$.
The budget, defined as the total number of individual instances that can be annotated, ranges from 10\% to 100\% with a stepsize of 10\%, where x\% means x\% of all instances in $\mcal{U}_0$ can be annotated.
%Therefore, we compare the curves of the mean $F_1$ vs. budget.

\subsection{Temporal Relation Extraction}
Temporal relations (TempRel) are a type of important relations representing the temporal ordering of events described by natural language text. That is to answer questions like which event happens earlier or later in time (see Fig.~\ref{fig:closure not working}a). 
% structure
Since time is physically one-dimensional, if \event{A} is before \event{B} and \event{B} is also before \event{C}, then \event{A} must be before \event{C}.
In practice, the label set for TempRels can be more complex, e.g., with labels such as \lbl{Simultaneous} and \lbl{Vague}, but the structure can still be represented by transitivity constraints (see Table~1 of \cite{NFWR18}), which can be viewed as an analogy of the chain structure in Example~\ref{ex:dag}.

% data
To avoid missing relations, annotators are required to exhaustively label every pair of events in a document (i.e., the complete annotation scheme), so it is necessary to study ESPA in this context.
Here we adopt the MATRES dataset \cite{NingWuRo18} 
%\footnote{\url{http://cogcomp.org/page/publication_view/834}} 
for its better inter-annotator agreement and relatively large size.
% We also assume that all gold events are given to avoid unnecessary complications.

Specifically, we use 35 documents as $\mcal{T}_0$ \QN{(the TimeBank-Dense section},\footnote{\QN{The original TimeBank-Dense section contains 36 documents, but in collecting MATRES, one of the documents was filtered out because it contained no TempRels between main-axis events.}} 147 documents as $\mcal{U}_0$ \QN{(the TimeBank section minus those documents in $\mcal{T}_0$)}, and the Platinum section (a benchmark testset of 20 documents with 1K TempRels) as $\mcal{T}_{test}$. Note that both schemes I and II are mimicked by downsampling the original annotations in MATRES, where the budget is defined as the total number of TempRels that are kept.
Following CogCompTime \cite{NZFPR18}, we choose the same features and sparse-averaged perceptron algorithm as the \textsc{Learn} component and ILP as \textsc{Inference} for SSPAN.

\begin{figure*}
	\centering
	\begin{subfigure}[b]{0.32\textwidth}
		\includegraphics[width=\textwidth]{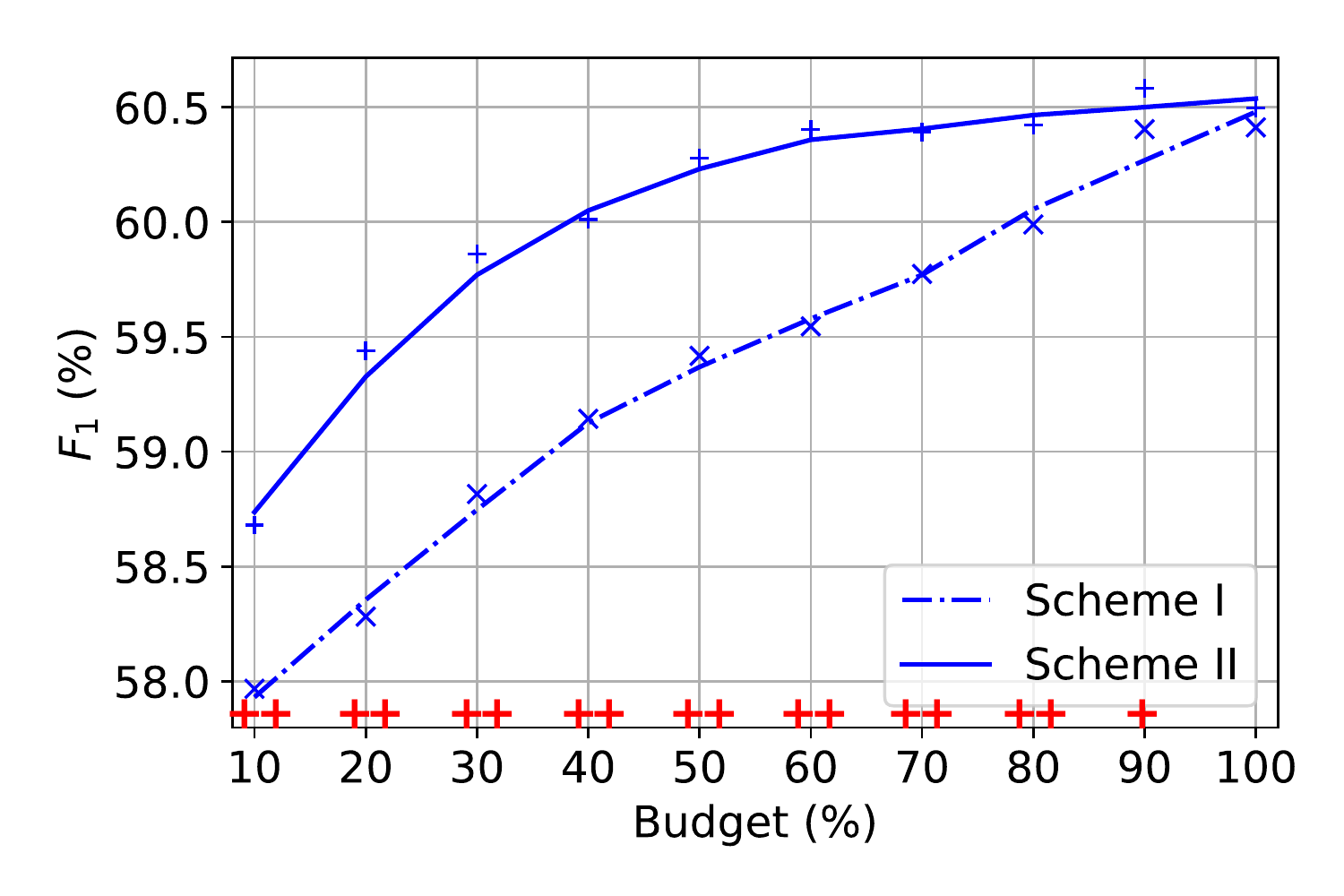}
		\caption{\small Temporal Relation Extraction}
		\label{fig:temopral result}
	\end{subfigure}
	\begin{subfigure}[b]{0.32\textwidth}
		\includegraphics[width=\textwidth]{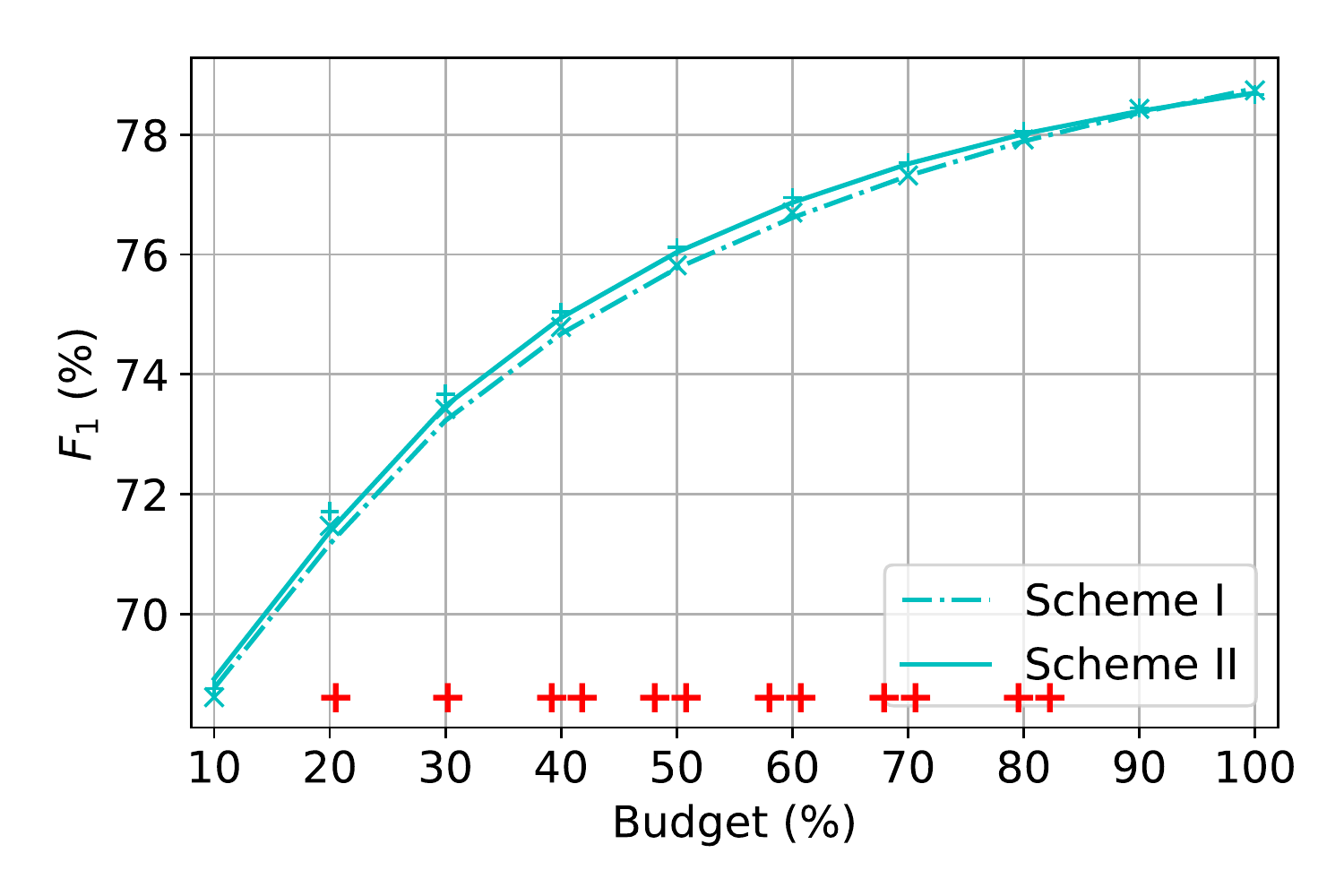}
		\caption{\small SRC}
		\label{fig:srl result}
	\end{subfigure}
	\begin{subfigure}[b]{0.32\textwidth}
		\includegraphics[width=\textwidth]{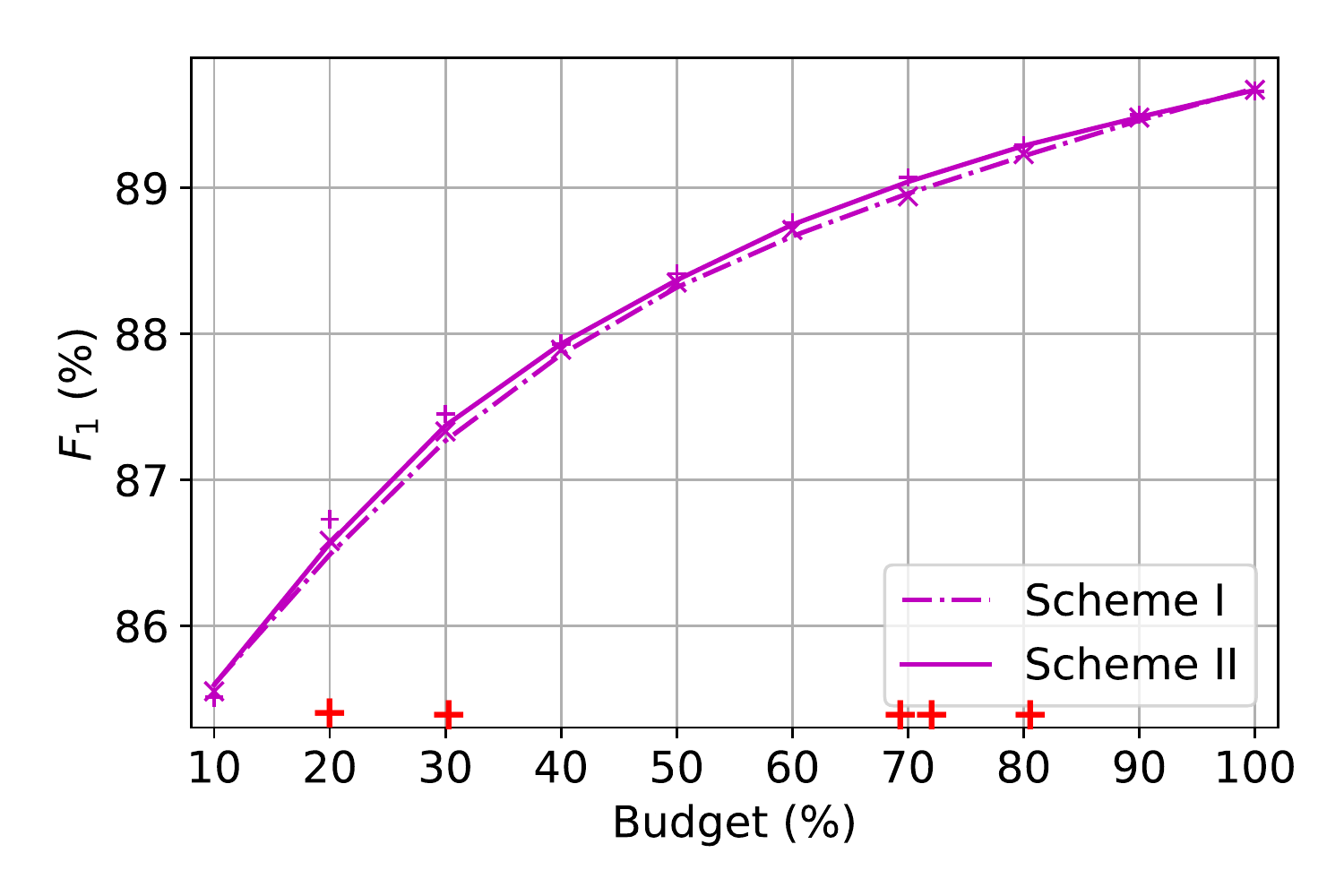}
		\caption{\small Shallow Parsing}
		\label{fig:chunker result}
	\end{subfigure}
	\caption{Comparison of the baseline, complete annotation scheme and the proposed ESPA scheme (See I \& II in Fig.~\ref{fig:comparison scheme}) under three structured learning tasks (note the scale difference). Each $F_1$ value is the average of 50 experiments, and each curve is based on corresponding $F_1$ values smoothed by Savitzky-Golay filters. We can see that \textbf{scheme II is consistently better than scheme I}. Per the Wilcoxon rank-sum test, the significance levels at each given budget are shown on the x-axes, where \textcolor{red}{$+$} and \textcolor{red}{$++$} mean $p<5\%$ and $p<1\%$, respectively.}\label{fig:result}
\end{figure*}

\subsection{Semantic Role Classification (SRC)}
Semantic role labeling (SRL) is to represent the semantic meanings of language and answer questions like {\em Who did What to Whom} and {\em When, Where, How} \cite{PalmerGiXu10}.
Semantic Role Classification (SRC) is a subtask of SRL, which assumes gold predicates and argument chunks and only classifies the semantic role of each argument. 
We use the Verb SRL dataset provided by the CoNLL-2005 shared task \cite{CarrerasMa05}, where the semantic roles include numbered arguments, e.g., \lbl{Arg0} and \lbl{Arg1}, and argument modifiers, e.g., location (\lbl{Am-Loc}), temporal (\lbl{Am-Tmp}), and manner (\lbl{Am-Mnr}) (see PropBank \cite{KingsburyPa02}).
The structural constraints for SRC is that each argument can be assigned to exactly one semantic role, and the same role cannot appear twice for a single verb,
%\footnote{except for \lbl{Am-Loc} and \lbl{Am-Mnr}, where this constraint holds for 99.9\% of the arguments \cite{Srikumar13}} 
so SRC is an assignment problem as in Example~\ref{ex:srl}.

Specifically, we use the Wall Street Journal (WSJ) part of Penn TreeBank  \uppercase\expandafter{\romannumeral3} \cite{MarcusSaMa93}.
We randomly select 700 sentences from the Sec.~24 of WSJ, among which 100 sentences as $\mcal{T}_0$ and 600 sentences as $\mcal{U}_0$. Our $\mcal{T}_{test}$ is 5700 sentences (about 40K arguments) from Secs.~00, 01, 23.
The budget here is defined as the total number of the arguments.
We adopt the SRL system in CogCompNLP \cite{KSZRCSRRLDTRMFWYSGUANLR18} and uses the sparse averaged perceptron as \textsc{Learn} and ILP as \textsc{Inference}.

% \missing{Hangfeng: Please follow Sec.4.1 and add descriptions here. (1) Some general descriptions about the dataset we use. (2) Specifically, what are $\mcal{T}_0$, $\mcal{U}_0$, and $\mcal{T}_{test}$, respectively. (3) What are the \textsc{Learn} and \textsc{Inference} here?}

\subsection{Shallow Parsing}
Shallow parsing, also referred as {\em chunking}, is a fundamental NLP task to identify constituents in a sentence, such as noun phrases (NP), verb phrases (VP), and adjective phrases (ADJP), which can be viewed as extending the standard BIO structure in Example~\ref{ex:chunking} with different chunk types: B-NP, I-NP, B-VP, I-VP, B-ADJP, I-ADJP, \dots, O.

We use the chunking dataset provided by the CoNLL-2000 shared task \cite{TjongBu00}. Specifically, we use 2K tokens' annotations as $\mcal{T}_0$, 14K tokens as $\mcal{U}_0$, and the benchmark testset (25K tokens) as $\mcal{T}_{test}$. The budget here is defined as the total number of tokens' BIO labels. 
The algorithm we use here is the chunker provided in CogCompNLP, where the \textsc{Learn} component is the sparse averaged perceptron and the \textsc{Inference} is described in \cite{PunyakanokRo01}.
%
% Note that randomly dropping annotations for shallow parsing leads to fractional chunks. For example, a sequence of $OBIIOOBIOO$ may turn into $\sqcap B\sqcap IO\sqcap\sqcap IOO$, which we admit is not realistic since annotators often label entire chunks at once. But here we are simply using shallow parsing to showcase the benefit of ESPA from the perspective of learning.

\subsection{Results}
We compare the $F_1$ performances of all three tasks in Fig.~\ref{fig:result}, averaged from 50 experiments with different randomizations.
As the budget increases, the system $F_1$ increases for both schemes I and II in all three tasks, which confirms the capability of the proposed SSPAN framework to learn from partial structures.
When the budget is 100\% (i.e., the entire $\mcal{U}_0$ is annotated), schemes I and II have negligible differences; when the budget is not large enough to cover the entire $\mcal{U}_0$, scheme II is consistently better than I in all tasks,
%(note the scale difference in subfigures), 
which follows our expectations based on the $I_k$ analysis.
{
% The improvement from scheme I to II is mostly statistically significant for each individual budget ratio; 
The strict improvement for all budget ratios indicates that the observation is definitely not by chance.}

\begin{figure}[h!]
	\centering\hspace{-1cm}
	\includegraphics[width=.5\textwidth]{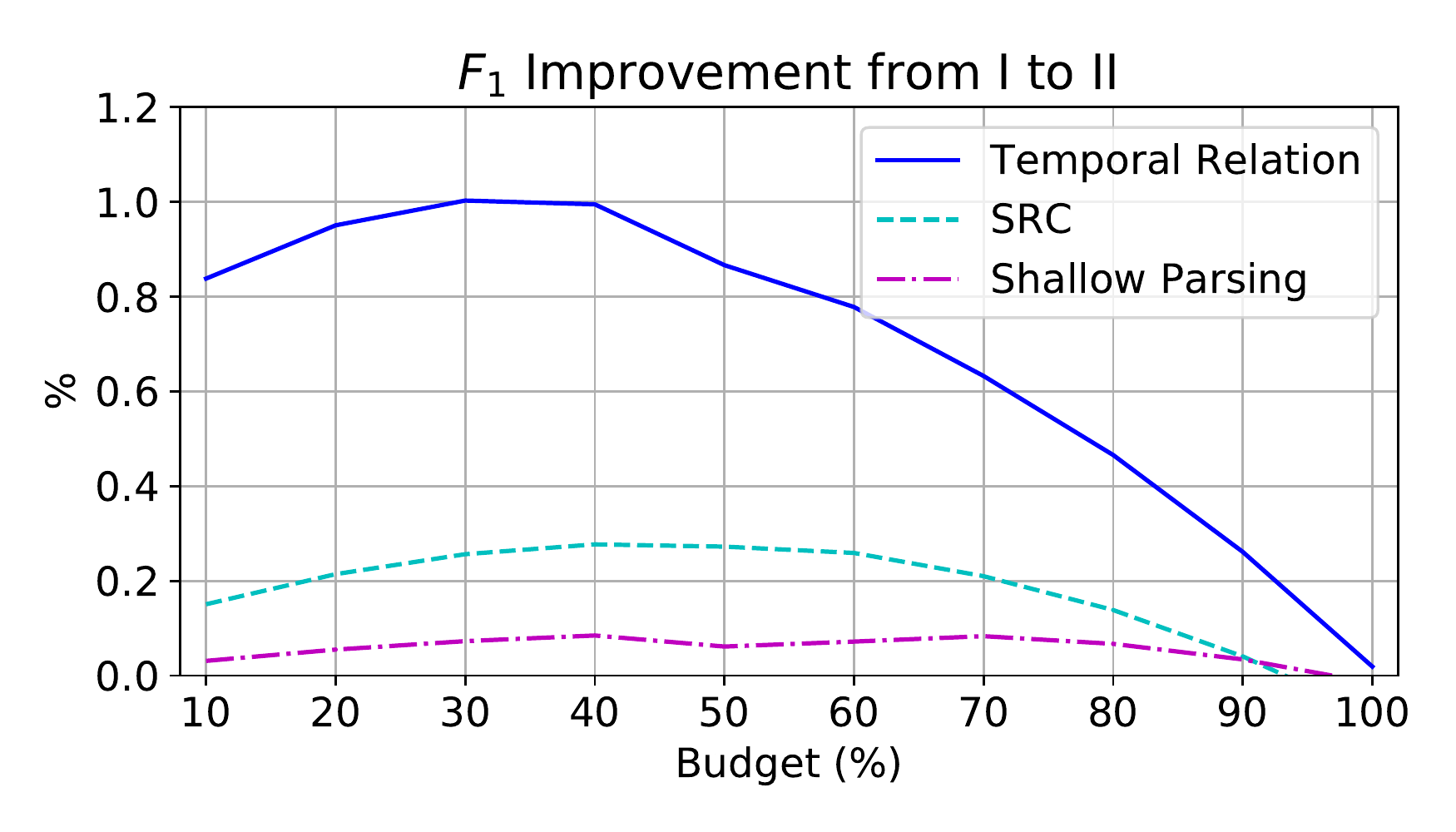}
	\caption{The improvement of $F_1$ brought by ESPA for each task in Fig.~\ref{fig:result}. Note that we conjectured earlier in Fig.~\ref{fig:compare convexity} that the BIO structure is the weakest among the three, which is consistent with the fact that shallow parsing benefits the least from ESPA.}
	\label{fig:improvement comparison}
\end{figure}

Figure~\ref{fig:improvement comparison} further compares the improvement from I to II across tasks.
When the budget goes down from 100\%, the advantage of ESPA is more prominent; but when the budget is too low, the quality of $\tilde{\mcal{P}}$ degrades and hurts the performance of SSPAN, leading to roughly hill-shaped curves in Fig.~\ref{fig:improvement comparison}.
We have also conjectured based on Fig.~\ref{fig:compare convexity} that the structure strength goes up from BIO chunks, to bipartite graphs, and to chains; interestingly, the improvement brought by ESPA is consistent with this order.
%However, TempRel extraction benefits more from ESPA than SRC does, which is unexpected according to Fig.~\ref{fig:compare convexity}. We think it is possibly due to two reasons. First, the analysis in Fig.~\ref{fig:compare convexity} is purely from the perspective of mutual information, which is not necessarily consistent with the system performances here in Fig.~\ref{fig:improvement comparison}.
%Second, the overall performance of SRC  (in the 80's) is already much higher than that of TempRel extraction (in the 60's), which apparently has less room for improvement, 

\QN{\textbf{Admittedly, the improvement, albeit statistically significant, is small, but it does not diminish
%dwarf 
the contribution of this paper}}: Our goal is to remind people that the ESPA scheme (or more generally, partialness) is, at the least, comparable to (or sometimes even better than) complete annotation schemes.
% prove that partial annotation schemes indeed have the capability to outperform complete schemes, and to what extent can we achieve the benefit of partialness is a future work we need to study. 
Also, the comparison here is in fact unfair to the partial scheme II, because we assume equal cost for both schemes, although it often costs less in a partial scheme as a large problem is decomposed into smaller parts.
Therefore, the results shown here implies that the information theoretical benefit of partialness can possibly offset its disadvantages for learning.

% SRC Overall Results:
% 0.1: 0.1455
% Full: 68.61
% Partial: 68.75
% Wilcoxon: 0.153
% Mann: 0.214

% 0.2: 0.2349
% Full: 71.47
% Partial: 71.71 
% Wilcoxon：0.021
% Mann: 0.037

% 0.3: 0.2471
% Wilcoxon: 0.024
% Mann: 0.020
% Full: 73.42
% Partial: 73.67

% 0.4: 0.2522
% Wilcoxon: 0.005
% Mann: 0.013
% Full: 74.79
% Partial: 75.04

% 0.5: 0.2996
% Wilcoxon: 0.001
% Mann: 0.0001
% Full: 75.82
% Partial: 76.12

% 0.6: 0.2494
% Wilcoxon: 0.0009
% Mann: 0.0008
% Full: 76.70
% Partial: 76.95

% 0.7: 0.2064
% Wilcoxon: 0.002
% Mann: 0.0008
% Full: 77.32
% Partial: 77.53

% 0.8: 0.1375
% WIlcoxon: 0.005
% Mann: 0.005
% Full: 77.92
% Partial: 78.06

% 0.9: 0.0219
% Wilcoxon: 0.39
% Mann: 0.42
% Full: 78.43
% Partial: 78.45

% 1.0: -0.0631
% Wilcoxon: 0.06
% Mann: 0.94
% Full: 78.74
% Partial: 78.67

% !TEX root = root.tex
\section{Dicussion and Conclusion}
\label{sec:discussion}
In this paper, we investigate a less studied, yet important question for structured learning: Given a limited annotation budget (either in time or money), which strategy is better, completely annotating each structure until the budget runs out, or annotating more structures at the cost of leaving some of them partially annotated?
\QN{\citet{NeubigMo10} investigated this issue specifically in annotating word boundaries and pronunciations for Japanese. Instead of annotating full sentences, they proposed to annotate only some words in a sentence (i.e., partially) that can be chosen heuristically (e.g., skip those that we have seen or those low frequency words). Conceptually, \citet{NeubigMo10} is an active learning work, with the understanding that if the order of annotation is deliberately designed, better learning can be achieved. The current paper addresses the problem from a different angle: Even without active learning, can we still answer the question above?}

The observation driving our questions is
%The motivation behind this is 
that when annotating a particular structure, the labels of the yet to be labeled variables may already be constrained by previous annotations and carry less information than those in a totally new structure.
Therefore, we systematically study the ESPA scheme -- stop annotating a given structure before it is completed and continue annotating another new structure.
%ESPA is rather convenient to implement either in crowdsourcing or not, and can be easily combined with active learning.

An important notion is annotation {\em cost}. Throughout the paper we have \QN{an ideal assumption} that the cost is linear in the total number of annotations, \QN{but in practice the case can be more complicated}.
\QN{First, the actual cost of each individual annotation may vary across different instances.} We try to eliminate this issue by enforcing random selection of annotation instances, rather than allowing the annotators to select arbitrarily by themselves. This strategy may be useful in practice as well, to avoid people only annotating easy cases.
\QN{Second, even if we only require labeling partial structures, it is likely that the annotator still needs to 
%read the sentence and 
comprehend the entire structure, incurring additional cost (usually in terms of time). This issue, however, is not addressed in this paper.}

% One caveat is that structural constraints may produce some annotations for free. For example, if, under transitivity constraints, we know from existing annotations that $A<B$ and $B<C$, then the annotation of $A<C$ is free.
% Since these free annotations exist no matter the annotation scheme is complete or partial, we conveniently assume that they incur the same cost as other instances.

Using this definition of cost, we provide a theoretical analysis for ESPA based on the mutual information between target structures and annotation processes. We show that for structures like chains, bipartite graphs, and BIO chunks, the information brought by an extra annotation attenuates as 
%a structure 
the annotation of the structure is more complete, suggesting to stop early and move to a new structure \QN{(although it still remains unclear when it is optimal to stop)}. 
This analysis is further supported by experiments on temporal relation extraction, semantic role classification, and shallow parsing, three tasks analogous to the three structures analyzed earlier, respectively.
The ratio of the attenuation curve as in Fig.~\ref{fig:compare convexity} is also shown to be an actionable metric to quantify the strength of a type of structure, which can be useful in various analysis, including judging whether ESPA is worthwhile for a particular task.
For example, a more detailed $I_k$-based analysis for SRC shows that predicates with more arguments are stronger structures than those with fewer arguments; we have investigated ESPA on those with more than 6 arguments and indeed, observed much larger improvement in SRC. More details on this analysis are put in the appendix.

\ignore{Note that we assume in this paper that when a structure is partial, we know perfectly which instances are annotated and which are not. We need to keep in mind that if this knowledge was absent, it would be more challenging to handle partialness. \dan{not clear why this comment is important}
}

\ignore{
The algorithms we use in those experiments are based on our SSPAN framework. There are possible improvements to it. For example, when learning from $\mcal{T}$ and $\tilde{\mcal{P}}$ (Line~\ref{ln:self training end} of Algorithm~\ref{algo:sspan}), we can keep those highly confident ones only and also put different weights to authentic labels and bootstrapped labels.
However, this paper focuses more on {\em whether} partial can be beneficial, rather than on {\em how} beneficial it is, and we leave it as a future work to develop more advanced algorithms to learn from partial annotations.
}

We think that the findings in this paper are very important.
First, as far as we know, we are the first to propose the mutual information analysis that provides a unique view of structured annotation, that of the reduction in the uncertainty of a target of interest $\mbf{Y}$ by another random variable/process. From this perspective, signals that have non-zero mutual information with $\mbf{Y}$ can be viewed as ``annotations''.  
These can be partially labeled structures (studied here), partial labels (restricting the possible labels rather than determining a single one  as in \QN{e.g., \citet{HLAR19}}, noisy labels (e.g., generated by crowdsourcing or heuristic rules) or, generally, other indirect supervision signals that are correlated with $\mbf{Y}$. As we proposed, these can be studied within our mutual information framework as well. This paper thus provides a way to analyze the benefit of general {\em incidental supervision signals} \cite{Roth17}) and possibly even provides guidance in selecting good incidental supervision signals.

Second, the findings here open up opportunities for new annotation schemes for structured learning. In the past, partially annotated training data have been either a compromise when completeness is infeasible (e.g., when ranking entries in gigantic databases), or collected freely without human annotators (e.g., based on heuristic rules).
If we intentionally ask human annotators for partial annotations, the annotation tasks can be more flexible and potentially, cost even less. This is because annotating complex structures typically require certain expertise,
and smaller tasks are often easier \cite{FernandesBr11}.
\QN{It is very likely that some complex annotation tasks require people to read dozens of pages of annotation guidelines, but once decomposed into smaller subtasks, even laymen can handle them.}
%\dan{I don't see the value of the NER example}
%For instance, when annotating named-entities, we can have specialized annotators for people, for locations, and for organizations, respectively, so that they can annotate faster and more reliably on their expertise. and distribute them randomly to a large corpus, without enforcing all the entities to be annotated.
Annotation schemes driven by crowdsourced question-answering, known to provide only partial coverage are successful examples of this idea \cite{HeLeZe15,MSHDZ17}.
Therefore, this paper is hopefully interesting to a broad audience.
% and can foster more investigations into the important problem of reducing the annotation cost of generating supervision signals for machine learning tasks.

%\HF{Should we add end-to-end SRL into the discussion to make it more complete?}

%without F as seed

%Partial: random sampling; biased sampling.

%UEM, PR. Some partial annotations and constraints may be soft, so PR is suitable.

%partialness: partial graph or partial label

\section*{Acknowledgements}
%\QN{
This research is supported in part by a grant from the Allen Institute for Artificial Intelligence (allenai.org); the IBM-ILLINOIS Center for Cognitive Computing Systems Research (C3SR) - a research collaboration as part of the IBM AI Horizons Network; Contract HR0011-15-2-0025 with the US Defense Advanced Research Projects Agency (DARPA); and by the Army Research Laboratory (ARL) and was accomplished under Cooperative Agreement Number W911NF-09-2-0053 (the ARL Network Science CTA).
The views and conclusions contained in this document are those of the authors and should not be interpreted as representing the official policies, either expressed or implied, of the Army Research Laboratory or the U.S. Government. The U.S. Government is authorized to reproduce and distribute reprints for Government purposes notwithstanding any copyright notation here on.
%}

\bibliography{ccg-long,cited-long}
\bibliographystyle{acl_natbib.bst}

\end{document}